\documentclass[runningheads]{llncs}
\usepackage{graphicx}

\usepackage{tikz}
\usepackage{comment}
\usepackage{amsmath,amssymb} %
\usepackage{color}
\usepackage{graphicx}
\usepackage{amsmath}
\usepackage{amssymb}
\usepackage{booktabs}
\usepackage{epsfig}
\usepackage{amsfonts}
\usepackage{url}
\usepackage{color}
\usepackage{bbm}
\usepackage{tabularx}
\usepackage{xspace}
\usepackage{booktabs}
\usepackage{microtype}
\usepackage{wrapfig}
\usepackage{adjustbox}
\usepackage{multirow}
\usepackage{afterpage}
\usepackage{makecell}
\usepackage{hyperref}

\usepackage[width=122mm,left=12mm,paperwidth=146mm,height=193mm,top=12mm,paperheight=217mm]{geometry}

\newcommand{\etal}{\textit{et al.}\xspace}

\newcommand{\WarpIt}{\tilde{I}_t}
\newcommand{\WarpDt}{\tilde{D}_t}

\newcommand{\RefineIt}{\hat{I}_t}
\newcommand{\RefineDt}{\hat{D}_t}

\newcommand{\RefineItm}{\hat{I}_{t-1}}
\newcommand{\RefineDtm}{\hat{D}_{t-1}}

\newcommand{\Tmax}{T_{\text{max}}}

\newcommand{\Lrecon}{\mathcal{L}_{\text{rec}}}
\newcommand{\Ladv}{\mathcal{L}_{\text{adv}}}

\usepackage{color}

\begin{document}
\pagestyle{headings}
\mainmatter
\def\ECCVSubNumber{2911}  %

\title{InfiniteNature-Zero: 
Learning Perpetual View Generation of Natural Scenes from Single Images} %

\titlerunning{InfiniteNature-Zero}
\author{Zhengqi Li$^{1}$ \and
Qianqian Wang$^{1, 2}$ \and 
Noah Snavely$^{1}$ \and
Angjoo Kanazawa$^{3}$}
\authorrunning{Li et al.}
\institute{Google Research \and
Cornell Tech, Cornell University \and
UC Berkeley}

\maketitle

\begin{abstract}
We present a method for learning 
to generate unbounded flythrough videos of natural scenes starting from a single view. This capability is learned from a collection of \emph{single photographs}, without requiring camera poses or even multiple views of each scene.
To achieve this, we propose a novel self-supervised 
view generation training paradigm where we sample and render virtual camera trajectories, including cyclic camera paths, allowing our model to learn stable view generation from a collection of single views.
At test time, despite never having seen a video, our approach can take a single image and generate long camera trajectories comprised of hundreds of new 
views with realistic and diverse content. We compare our approach with recent state-of-the-art supervised view generation methods that require posed multi-view videos 
and demonstrate superior performance and synthesis quality.
Our project webpage, including video results, is at \url{infinite-nature-zero.github.io}.

\end{abstract}

\section{Introduction}
\begin{figure}[t]
\centering
  \includegraphics[width=0.95\columnwidth]{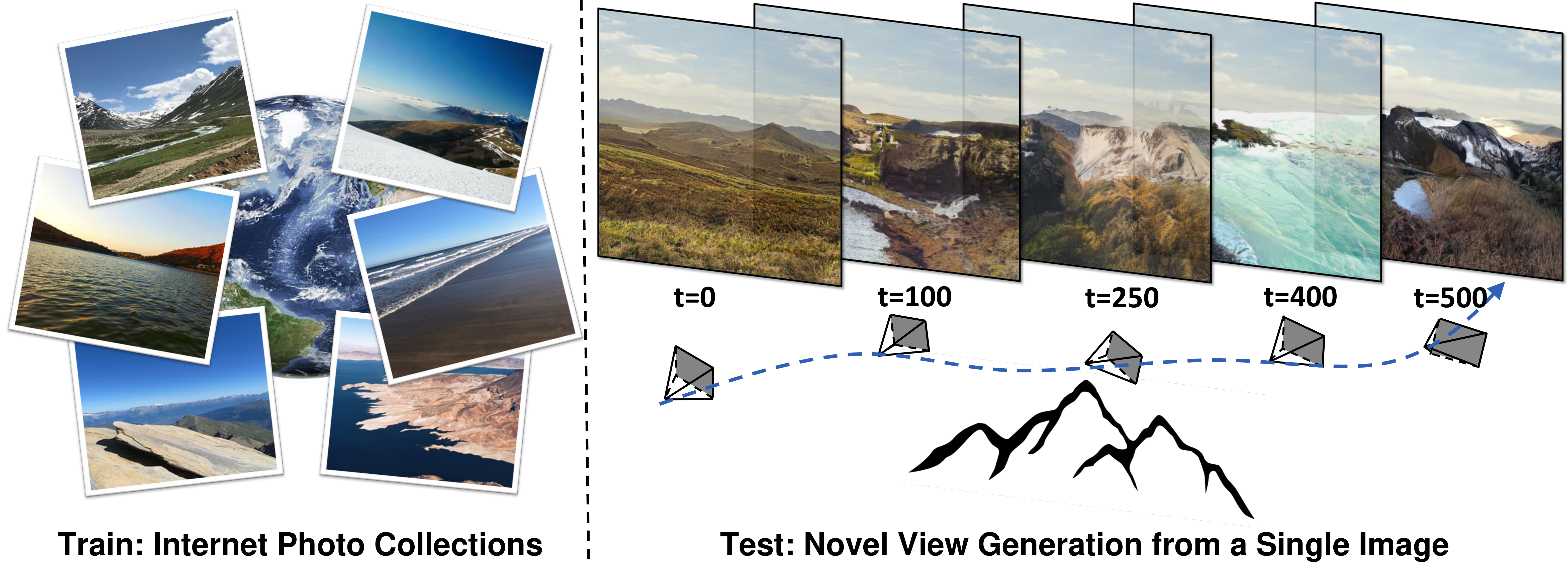}
  \caption{\textbf{Learning perpetual view generation from single images.} Given a single RGB image input, our approach generates novel views corresponding to a continuous long camera trajectory, without ever seeing a video during training.
  }
\label{fig:teaser}
\end{figure}

There are millions of photos of natural landscapes on the Internet, capturing 
breathtaking scenery across the world. 
Recent advances in vision and graphics have led to the ability to turn such images into compelling 3D photos~\cite{kopf2020one,shih20203d,jampani2021slide}. 
However, most prior work can only extrapolate scene content within a limited range of views corresponding to a small head movement. What if, instead, we could step into the picture and fly through the scene like a bird and explore the world in 3D,
and see diverse elements like mountain, lakes, and forests appear naturally as we move through the landscape? 
This challenging new task was recently proposed by Liu~\etal~\cite{liu2021infinite}, who called it \emph{perpetual view generation}: given a single RGB image, the goal is to synthesize a video depicting a scene captured from a moving camera with an arbitrary long camera trajectory. Methods that tackle this problem have applications in content creation and virtual reality.

However, perceptual view generation is a very challenging problem: as the camera travels through the world, we must fill in unseen missing regions in a harmonious manner, and must resolve new details as scene content approaches the camera, all the while maintaining photo-realism and diversity.
Liu~\etal~\cite{liu2021infinite} 
proposed a supervised solution that generates view sequences in an auto-regressive manner. To train their model, 
Liu~\etal (which we will refer to as \emph{Infinite Nature}), require a large dataset of video clips of nature scenes along with per-frame camera poses.
In essence, perpetual view generation is a video synthesis task, but the requirement of \emph{posed} video makes data collection very challenging. Obtaining large amounts of diverse, high-quality, and long videos of nature scenes is difficult enough, let alone estimating accurate camera poses on these videos at scale.
In contrast, Internet photos of nature landscapes are much easier to collect, and have spurred 
research 
into panorama synthesis~\cite{skorokhodov2021aligning,lin2021infinity}, image extrapolation~\cite{InOut,saharia2021palette}, image editing~\cite{park2020swapping}, and multi-model image synthesis~\cite{esser2021taming,huang2021multimodal}. %

Can we use existing \emph{single}-image datasets for perpetual 3D view generation?
In other words, can we learn view generation by simply observing many photos, without requiring video or camera poses? 
Training with less powerful supervision would seemingly make this already challenging synthesis task even harder. 
And doing so is not a straightforward application of prior methods. For instance, prior single-image view synthesis methods either require posed multi-view data~\cite{wiles2020synsin,rockwell2021pixelsynth,koh2021pathdreamer}, or can only extrapolate within a limited range of viewpoints~\cite{kopf2020one,shih20203d,jampani2021slide,hu2021worldsheet}.
Other methods for video synthesis~\cite{Akan2021ICCV,tulyakov2018mocogan,yu2022digan,lee2021revisiting} require 
videos spanning multiple views as training data, and can only generate a limited number of novel frames with no ability to control camera motion at runtime.

In this work, we present a new method for learning perpetual view generation from only a collection of single photos, without requiring multiple views of each scene or camera information. 
Despite using much less information, our approach improves upon the visual quality of prior methods that require multi-view data.
We do so by utilizing \emph{virtual} camera trajectories and computing
losses that enable high-quality perpetual view generation results. 
Specifically, we first introduce a self-supervised view synthesis strategy that utilizes \emph{cyclic} virtual camera trajectories, where we know that the synthesized end frame should be identical to the starting frame. 
This idea provides the network a training signal for generating a single view synthesis step without multi-view data.
Second, to learn to
generate a long sequence of novel views we employ an adversarial perpetual view generation training technique, encouraging views along a long virtual camera trajectory to be realistic and generation to be stable. The only requirement for our approach is an off-the-shelf monocular depth network to obtain disparity for the initial frame, but this depth network does not need to be trained on our data. In this sense, our method is self-supervised, leveraging underlying pixel statistics from single-image collections. %
Because we train with no video data whatsoever, we call our approach \emph{InfiniteNature-Zero}.

We show that training our model using prior video/view generation methods leads to training divergence or mode collapse. We therefore introduce balanced GAN sampling and progressive trajectory growing strategies that stabilize model training. In addition, to prevent artifacts and drift during inference, we propose a global sky correction technique that yields more consistent and realistic synthesis results along long camera trajectories.

We evaluate our
method on two public nature scene datasets, and compare to recent supervised video synthesis and view generation methods. We demonstrate superior performance compared to state-of-the-art baselines trained on multi-view collections, even though our model only requires single-view photos during training. To our knowledge, our work is the first to tackle unbounded 3D view generation for natural scenes trained on 2D image collections, 
and believe this capability will enable new methods for generative 3D synthesis that leverage more limited supervision.

\section{Related Work}

\noindent \textbf{Image extrapolation.}
An inspiring early approach to infinite view extrapolation, called \emph{Infinite Images} was proposed by Kaneva~\etal~\cite{Kaneva_2010}, which continually retrieves, transforms, and blends 
imagery from a database to create an infinite 2D landscape. We revisit this idea in a 3D context, which requires 
inpainting, i.e., filling missing content within an image \cite{hays2007scene,yu2018generative,yu2019free,liu2021pd,zhao2021large}, as well as \emph{outpainting}, extending the image and inferring unseen content outside the image boundaries~\cite{wang2019wide,yang2019very,teterwak2019boundless,bowen2021oconet,rockwell2021pixelsynth,saharia2021palette} in order to generate images from novel camera viewpoints. 
Super-resolution~\cite{glasner2009super,ledig2017photo} is also an important 
aspect of perpetual view generation, as approaching a distant object requires synthesizing additional high-resolution detail.
Image-specific GAN methods demonstrate super-resolution of textures and natural images as a form of image extrapolation \cite{zhou2018non,shocher2018zero,shaham2019singan,shocher2018ingan}.
In contrast to the above methods that address these problems individually, our methods handles inpainting, outpainting, and superresolution jointly.

\smallskip
\noindent \textbf{Generative view synthesis.}
View synthesis is the problem of generating novel views of a scene from existing views. Many view synthesis methods require multiple views of a scene as input~\cite{levoy1996light,chaurasia2013depth,zhou2018stereo,mildenhall2019local,flynn2019deepview,extremeview,lombardi2019neural,Riegler2020FVS,mildenhall2020nerf,wang2021ibrnet,liu2020neural}, though recent works can also generate novel views from a single image~\cite{chen2019mono,tulsiani2018layer,niklaus20193d,single_view_mpi,shi2014light,wiles2020synsin,jang2021codenerf,Shih3DP20,Kopf-OneShot-2020,rombach2021geometry}. These methods often require multi-view posed datasets such as RealEstate10k~\cite{zhou2018stereo}. 
However, empowered by advances in neural rendering, recent works show that one can unconditionally generate 3D scene representations for 3D-consistent image synthesis~\cite{HoloGAN2019,schwarz2020graf,niemeyer2021giraffe,devries2021unconstrained,niemeyer2021campari,gu2021stylenerf,Chan2021}. Many of these methods only require unstructured 2D images for training.
When GAN inversion is possible, these methods can also be used for single-image view synthesis, although they have only been demonstrated on specific object categories like faces~\cite{chan2021pi,Chan2021}. All of the works above only allow for a limited range of output viewpoints.
In contrast, our method can generate new views perpetually, eventually reaching an entirely new distant view, from a single input image. Most related to our work is Liu \etal~\cite{liu2021infinite}, which also performs perpetual view generation. However, Liu \etal require posed videos during training. Our method can be trained with unstructured 2D images, and also experimentally achieves better view generation diversity and quality.

\smallskip
\noindent \textbf{Video synthesis.}
Our work is also related to video synthesis~\cite{clark2019adversarial,tian2021good}, which can be roughly divided into three categories: 
1) unconditional video generation~\cite{tulyakov2018mocogan,munoz2021temporal,fox2021stylevideogan,liu2021content}, which produces a video sequence from an input noise; 
2) video prediction~\cite{vondrick2016generating,vondrick2017generating,wang2017predrnn,villegas2017decomposing,hsieh2018learning,lee2021revisiting}, which generates a video sequence from one or more initial observations; and 
3) video-to-video synthesis, which maps a video from a source domain to a target domain. Most video prediction methods focus on generating videos of dynamic objects under a static camera~\cite{vondrick2016generating,finn2016unsupervised,vondrick2017generating,denton2018stochastic,ye2019cvp,yu2022generating,lee2021revisiting}, e.g., human motion~\cite{blank2005actions} or the movement of robot arms~\cite{finn2016unsupervised}. 
In contrast, we focus on generating new views of static nature scenes with a moving camera. Several video prediction methods can also simulate moving cameras~\cite{clark2019efficient,villegas2019high,Akan2021ICCV,lee2021revisiting}, but unlike our approach, they require long video sequences for training, do not reason about underlying 3D scene geometry, and do not allow for explicit control over camera viewpoint. %
More recently, Koh \etal propose to navigate and synthesize indoor scenes with controllable camera motion~\cite{koh2021pathdreamer}. 
However, they require ground truth RGBD panoramas as supervision %
and can only generate novel frames up to 6 steps. Many prior methods in this vein also require 3D inputs, such as voxel grids~\cite{hao2021gancraft} or dense point clouds~\cite{mallya2020world}, whereas we require only a single RGB image.

\section{Learning view generation from single-image collections}

We formulate the task of perpetual view generation as follows: given an starting RGB image $I_0$, 
generate an image sequence $(\hat{I_1}, \hat{I_2}, ..., \hat{I_t},...)$ corresponding to an arbitrary camera trajectory $(c_1, c_2, ..., c_t, ...)$ starting from $I_0$, where the camera viewpoints ${c_t}$ can be specified either algorithmically or via user input. 

The prior Infinite Nature method tackles this problem by decomposing it into three phases:
\textbf{render}, \textbf{refine} and \textbf{repeat}~\cite{liu2021infinite}. Given an RGBD image $(\RefineItm, \RefineDtm)$ at camera $c_{t-1}$, the \textbf{render} phase renders a new view $(\WarpIt, \WarpDt)$ at $c_{t}$ by transforming and warping $(\RefineItm, \RefineDtm)$ using a differentiable 3D renderer $\mathcal{W}$. This yields a warped view $(\WarpIt, \WarpDt) = \mathcal{W} \left( (I_{t-1}, D_{t-1}), T_{t-1}^{t} \right)$, 
where $T_{t-1}^{t}$ is an $SE(3)$ transformation from $c_{t-1}$ to $c_t$. In the \textbf{refine} phase, the warped RGBD image $(\WarpIt, \WarpDt)$ is fed into a refinement network $F_{\theta}$ to fill in missing content and add details: $(\RefineIt, \RefineDt) = F_{\theta}(\WarpIt, \WarpDt)$. The refined outputs $(\RefineIt, \RefineDt)$ are then treated as a starting view for the next iteration of the \textbf{repeat} step, from which the process iterates.
We refer readers to the original work for more details~\cite{liu2021infinite}.

To supervise a view generation model, Infinite Nature trains on video clips of natural scenes, where each video frame has camera pose dervied from structure from motion (SfM)~\cite{zhou2018stereo}. During training, it randomly chooses one frame in a video clip as the starting view $I_0$, and performs the render-refine-repeat process along the provided SfM camera trajectory. At a camera viewpoint $c_t$ along the trajectory, a reconstruction loss and an adversarial loss are computed between the image predicted by the network $(\RefineIt , \RefineDt)$ and the corresponding real RGBD frame $(I_t, D_t)$. However, obtaining long nature videos with accurate camera poses is difficult due to often distant or non-Lambertian contents of landscape scenes (e.g., sea, mountain, and sky). In contrast, our method does not require videos at all, whether with camera poses or not. 

\begin{figure}[t]
\centering
  \includegraphics[width=0.95\columnwidth]{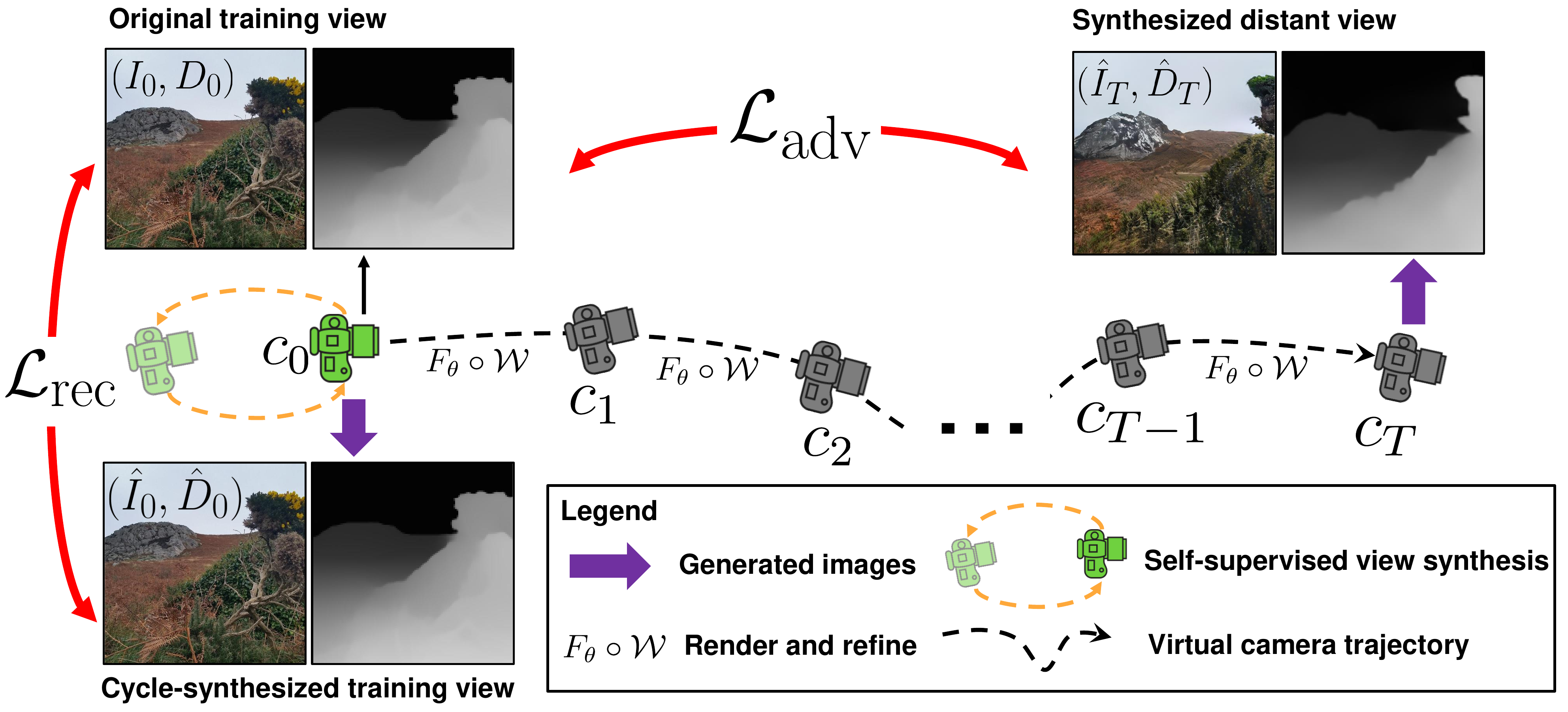} %
  \caption{\textbf{Self-supervised view generation via virtual cameras.}
  Given a starting RGBD image $(I_0, D_0)$ at viewpoint $c_0$, our training procedure samples two virtual camera trajectories: 1) a cycle to and back from a single virtual view (dashed orange arrows), creating a \emph{self-supervised view synthesis} signal enforced by the reconstruction loss $\Lrecon$.
  2) a longer virtual camera path for which we generate corresponding images via the render-refine-repeat process (black dashed arrows and gray cameras). An adversarial loss $\Ladv$ between the final view $(\hat{I}_T, \hat{D}_T)$ and the real image $(I_0, D_0)$ enables the network to learn long-range view generation.
  }  
\label{fig:overview_1}
\end{figure}

We show that 2D photo collections alone provide sufficient supervision signals to learn perceptual view generation, given an off-the-shelf monocular depth prediction network.
Our key idea is to sample and render \emph{virtual} camera trajectories starting from the training image, using the refined depth at each frame to warp it to the next view. 
We generate two kinds of camera trajectories, illustrated in Fig.~\ref{fig:overview_1}: 
First, we produce \emph{cyclic} camera trajectories that start and end at the training image. Since the start and end frame should be identical, we can use a reconstruction loss on the initial frame as a self-supervised loss (Sec.~\ref{sec:self_sup_view_syn}). 
This self-supervision trains our network to do 
geometry-aware view refinement during view generation.
Second, we synthesize longer virtual camera paths and compute an adversarial loss $\Ladv$ on the final rendered image (Sec.~\ref{sec:training_view_generation}). 
This signal trains our network to learn stable view generation over long camera trajectories. The rest of this section describes the two training signals in detail, as well as a sky correction component (Sec.~\ref{sec:sky_correction}) that prevents drift in sky regions at test time, yielding more realistic and stable long-range trajectories for nature scenes.

\subsection{Self-supervised view synthesis} 
\label{sec:self_sup_view_syn}

\begin{figure}[t]
\centering
  \includegraphics[width=0.95\columnwidth]{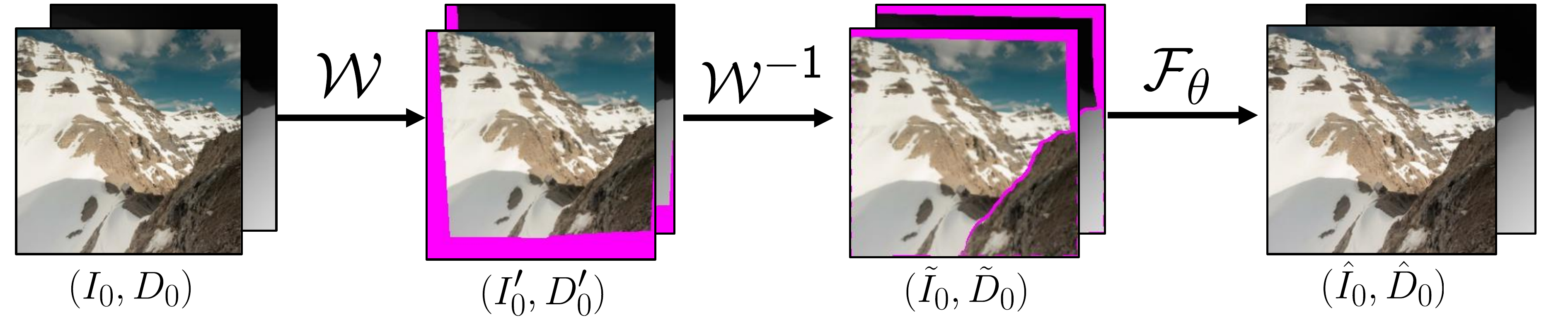} %
  \caption{\textbf{Self-supervised view synthesis.} From a real RGBD image $(I_0, D_0)$, we synthesize an input $(\tilde{I}_0, \tilde{D}_0)$ to a refinement model by cycle-rendering through a virtual viewpoint.
  From left to right: input image; input rendered to a virtual ``previous'' view; virtual view rendered \emph{back} to the starting viewpoint; final image $(\hat{I}_0, \hat{D}_0)$ refined with refinement network $\mathcal{F}_\theta$,  trained to match the starting image. %
  }
\label{fig:double_splatting}
\end{figure}

In Infinite Nature's supervised learning framework, a reconstruction loss is applied between predicted and corresponding real RGBD images to train the network to refine the inputs rendered from a previous viewpoint. 
Note that unlike the task of free-form image inpainting~\cite{zhao2021large}, this next-view supervision provides a crucial signal for the network to learn to add suitable details and to fill in missing regions around disocclusions using background context, while preserving 3D perspective. Accordingly, we cannot fully simulate the necessary 3D training signals from standard 2D inpainting supervision alone.
Instead, our idea is to treat the known real image as the held-out ``next'' view, and simulate a rendered image input from a virtual ``previous'' viewpoint. We implement this idea 
by rendering a \emph{cyclic} virtual camera trajectory starting and ending at the known input training view, then comparing the final rendered image at the end of the cycle to the known ground truth input view. In practice, we find that a cycle including just one other virtual view (i.e., warping to a sampled viewpoint, then rendering back to the input viewpoint) is sufficient. Fig.~\ref{fig:double_splatting} shows an example sequence of views produced in such a cyclic rendering step.

To implement this idea, we first predict the depth $D_0$ from a real image $I_0$ using a 
standard monocular depth network~\cite{ranftl2020towards}.
We randomly sample a nearby viewpoint with relative camera pose $T$ within a set of maximum values for each camera parameter. 
We then synthesize the view at virtual pose $T$ by rendering $(I_0, D_0)$ to a new image $(I'_0, D'_0) = \mathcal{W} \left( (I_0, D_0), T \right)$. 
Next, to encourage the network to learn to fill in missing 
content at disocclusions, we create a per-pixel binary mask $M'_0$ derived from the rendered depth $D'_0$ at the virtual viewpoint~\cite{liu2021infinite,jampani2021slide}. 
Finally, we render this virtual view with mask $(I'_0, D'_0, M'_0)$ back to the starting viewpoint via transform $T^{-1}$: $(\tilde{I}_0, \tilde{D}_0, \tilde{M}_0) = \mathcal{W} \left( (I'_0, D'_0, M'_0), T^{-1}\right)$ where the rendered mask is element-wise multiplied with the rendered RGBD image. Intuitively, this strategy constructs inputs whose pixel statistics, including blur and missing content, are similar to those produced by warping a view forward to a next viewpoint, yielding naturalistic input to view refinement. %

The cycle-rendered images $(\tilde{I}_0, \tilde{D}_0)$ are then fed into the refinement network $F_{\theta}$, whose outputs $(\hat{I}_0, \hat{D}_0) = F_{\theta} (\tilde{I}_0, \tilde{D}_0)$ are compared to the original RGBD image $(I_0, D_0)$ to yield a reconstruction loss $\Lrecon$.
Because this method does not require actual multiple views or SfM camera poses, we can generate an effectively infinite set of virtual camera motions during training. Because the target view is always an input training view we seek to reconstruct, this approach can be thought of as a self-supervised way of training view synthesis.

\subsection{Adversarial perpetual view generation} \label{sec:training_view_generation}
Although the insight above enables the network to learn to refine a rendered image,
directly applying such a network iteratively during inference over multiple steps %
quickly degenerates (see third row of Fig.~\ref{fig:ablation}). As observed by prior work~\cite{liu2021infinite}, we must train a synthesis model through multiple recurrently-generated camera viewpoints in order for the view generation to be stable.
Therefore, 
in addition to the self-supervised training in Sec.~\ref{sec:self_sup_view_syn}, we also train on longer virtual camera trajectories.
In particular, during training, for a given input RGBD image $(I_0, D_0)$, we randomly sample a virtual camera trajectory 
$\left( c_1, c_2, ..., c_T \right)$ starting from $(I_0, D_0)$ by iteratively performing render-refine-repeat $T$ times, yielding 
a sequence of generated views $(\hat{I}_1, \hat{I}_2, ..., \hat{I}_T)$.
To prevent the camera from traversing out-of-distribution viewpoints (e.g., crashing into mountains or water)
we adopt the auto-pilot algorithm from~\cite{liu2021infinite} to sample the camera path.
The auto-pilot algorithm determines the pose of the next view based on the proportion of sky and foreground elements as determined by the estimated disparity map at the current viewpoint (see supplemental material for more details). Next, we discuss how we train our model using such sampled virtual camera trajectories. 

\smallskip
\noindent \textbf{Balanced GAN sampling.}
We now have a generated sequence of views along a virtual camera trajectory from the input image, but we do not have the ground truth sequence corresponding to these views. 
How can we train the model without such ground truth? We find that it is 
sufficient to compute an adversarial loss that trains a discriminator to distinguish between
real images and
the synthesized ``fake'' images along the virtual camera path. 
One straightforward implementation of this idea is %
to treat all $T$ predictions $\{\RefineIt, \RefineDt\}_{t=1}^T$, along the virtual path as fake samples, and sample $T$ real images randomly from the dataset. %
However, this strategy leads to unstable training, because there is a significant discrepancy in pixel statistics between the generated view sequence and the set of sampled real photos: a generated sequence along a camera trajectory has frames with similar content with smoothly changing viewpoints, whereas randomly sampled real images from the dataset exhibit completely different content and viewpoints. This vast difference in the distribution of images that the discriminator observes leads to unstable training in conditional GAN settings~\cite{hao2021gancraft}.
To address this issue, we propose a simple but effective technique to stabilize the training. Specifically, for a generated sequence, we only feed the discriminator the generated image $(\hat{I}_T, \hat{D}_T)$ at the last camera $c_T$ as the fake sample, and use its corresponding input image $(I_0, D_0)$ at the starting view as the real sample, as shown in Fig.~\ref{fig:overview_1}. In this case, the real and fake sample in each batch will exhibit similar content and viewpoint variations.
Further, during each training iteration, we randomly sample the length of virtual camera trajectory $T$ between $1$ and a predefined maximum length $\Tmax$, so that the prediction at any viewpoint and step will be sufficiently trained.

\smallskip
\noindent \textbf{Progressive trajectory growing.}
We observe that without the guidance of ground truth sequences,
the discriminator quickly gains an overwhelming advantage over the generator at the beginning of training. 
Similarly to issues explored in prior work on 2D GANs~\cite{Karras2018ProgressiveGO,karnewar2020msg,Shaham2019SinGANLA}, we find that it takes longer for the network to predict plausible views at more distant viewpoints. As a result, the discriminator will easily distinguish real images from fake ones generated at distant views, 
and hence offer meaningless gradients to the generator. To address this issue, we propose to progressively grow the length of the virtual camera trajectory.
We begin with self-supervised view synthesis as described in Sec.~\ref{sec:self_sup_view_syn} and pretrain the model for 200K steps. 
We then increase the maximum length of the virtual camera trajectory $T$ 
by 1 every 25K iterations until reaching a predefined maximum length $\Tmax$.  This progressive growing strategy ensures that images rendered at a previous viewpoint $c_{t-1}$ have been sufficiently initialized before being fed to the refinement network to generate the view at the next viewpoint $c_{t}$.

\subsection{Global sky correction}
\label{sec:sky_correction}
The sky is an indispensable visual element  
of nature scenes with unique characteristics---it should change much more slowly than the foreground content, since the sky is at infinity. 
However, we found that the sky synthesized by Infinite Nature can contain unrealistic artifacts after multiple steps. We also found that monocular depth predictions can be inaccurate in sky regions, leading to sky contents to quickly approach the camera in an unrealistic manner.

Therefore, we devise a method to correct the sky regions of refined RGBD images at each test time iteration by leveraging the sky content from the starting view. 
In particular, we use an off-the-shelf semantic segmentation method~\cite{chen2017rethinking} and the predicted disparity map to determine soft sky masks for the starting and for each generated view, which we found to be effective in identifying sky pixels. We then correct the sky texture and disparity at every step by alpha blending the homography-warped sky content from the starting view (warped according to the camera rotation's effect on the plane at infinity) with the foreground content in the current generated view. 
To avoid redundantly outpainting the same sky regions, we expand the input image and disparity through GAN inversion~\cite{chong2021stylegan,InOut} to seamlessly create a canvas of higher resolution and field of view. We refer readers to the supplemental material for more details.
As shown in the penultimate column of Fig.~\ref{fig:ablation}, when applying global sky correction at test time, sky regions exhibit significantly fewer artifacts, resulting in more realistic generated views. %

\subsection{Network and supervision losses}
\label{sec:loss}
We adopt a variant of the CoMod-GAN conditional StyleGAN model~\cite{zhao2021large} as our backbone refinement module $F_{\theta}$. 
Specifically, $F_\theta$ consists of 
a global encoder and a StyleGAN generator, where the encoder produces a global latent code $z_0$ from the input view. At each refine step, we co-modulate intermediate feature layers of the StyleGAN generator via concatenation of $z_{0}$ and a latent code $z$ mapped from Gaussian noise.
The training loss for the generator and discriminator is:
\begin{align}
    \mathcal{L}^F = \mathcal{L}^F_{\text{adv}} + \lambda_1 \Lrecon, \quad \mathcal{L}^D  = \mathcal{L}^D_{\text{adv}} +  \lambda_2 \mathcal{L}_{R_1}
\end{align}
where $\mathcal{L}^F_{\text{adv}}$ and $ \mathcal{L}^D_{\text{adv}}$ are non-saturated GAN losses~\cite{goodfellow2014generative}, applied on the last view from the camera trajectory and the corresponding training image.
$\mathcal{L}_{\text{rec}}$ is a reconstruction loss between real images (and depth maps) and their corresponding cycle-synthesized views described in Sec~\ref{sec:self_sup_view_syn}: 
$\mathcal{L}_{\text{rec}} =  \sum_l || \phi^l(\hat{I}_0) - \phi^l(I_0)||_1 + || \hat{D}_0 - D_0||_1$, where $\phi^{l}$ is a feature map 
at scale $l$ from different layers of a pretrained VGG network~\cite{sengupta2019going}. $\mathcal{L}_{R_1}$ is a gradient regularization term that is applied to the discriminator during training~\cite{karras2020analyzing}.

\section{Experiments}

\begin{figure}[t]
    \centering
    \setlength{\tabcolsep}{0.01cm}
    \renewcommand{\arraystretch}{0.5}
    \begin{tabular}{ccccccc}
        \includegraphics[width=0.14\columnwidth]{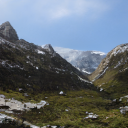} & 
        \includegraphics[width=0.14\columnwidth]{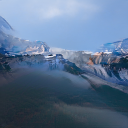} &
        \includegraphics[width=0.14\columnwidth]{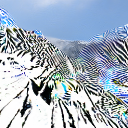} & 
        \includegraphics[width=0.14\columnwidth]{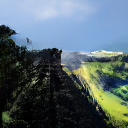} & 
        \includegraphics[width=0.14\columnwidth]{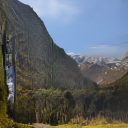} & 
        \includegraphics[width=0.14\columnwidth]{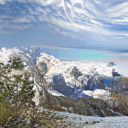} &
        \includegraphics[width=0.14\columnwidth]{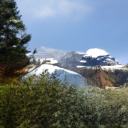} 
        \\
        \includegraphics[width=0.14\columnwidth]{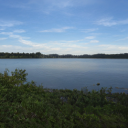} & 
        \includegraphics[width=0.14\columnwidth]{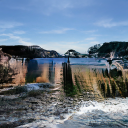} &
        \includegraphics[width=0.14\columnwidth]{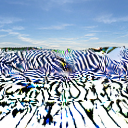} & 
        \includegraphics[width=0.14\columnwidth]{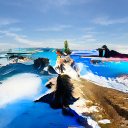} & 
        \includegraphics[width=0.14\columnwidth]{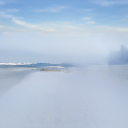} & 
        \includegraphics[width=0.14\columnwidth]{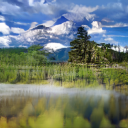} &
        \includegraphics[width=0.14\columnwidth]{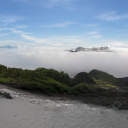}
        \\
        \vphantom{I}\footnotesize $I_0$ &
        \vphantom{I}\footnotesize w/o BGS &
        \vphantom{I}\footnotesize w/o repeat &
        \vphantom{I}\footnotesize w/o PTG &
        \vphantom{I}\footnotesize w/o SVS &
        \vphantom{I}\footnotesize w/o sky  &
        \vphantom{I}\footnotesize Full 
    \end{tabular}
    \caption{\textbf{Generated views after 50 steps with different settings.} Each row shows results for a different input image. From left to right: input view; results without balanced GAN sampling; without the adversarial perpetual view generation strategy; without progressive trajectory growing; without self-supervised view synthesis; without global sky correction; full approach.}
    \label{fig:ablation}
\end{figure}

\subsection{Datasets and baselines} \label{sec:baseline}
We evaluate our approach on two public datasets of nature scenes: the Landscape High Quality (LHQ) dataset~\cite{skorokhodov2021aligning}, a collection of 90K landscapes photos collected from the Internet, and the Aerial Coastline Imagery Dataset (ACID)~\cite{liu2021infinite}, a video dataset of nature scenes with SfM camera poses.

On the ACID dataset, where posed video data is available, we compare with several state-of-the-art supervised learning methods. 
Our main baseline is Infinite Nature, the recent state-of-the-art view generation method designed for natural scenes~\cite{liu2021infinite}. 
We also compare with other recent view and video synthesis methods, including geometry-free view synthesis (GFVS)~\cite{rombach2021geometry} and PixelSynth~\cite{rockwell2021pixelsynth}, both of which are based on VQ-VAE~\cite{razavi2019generating,esser2021taming} for long-range view synthesis.
Additionally, we compare with two recent video synthesis methods, SLAMP~\cite{Akan2021ICCV} and DIGAN~\cite{yu2022digan}. 
Following their original protocols, we train both methods with video clips of 16 frames from the ACID dataset until convergence. 

For the LHQ dataset, since there is no multi-view training data and we are unaware of prior methods that can train on single images, we show results from our approach with different configurations, described in more detail in Sec.~\ref{sec:ablations}.

\newcommand{\tablespace}{\,\,\,\,}
\newcommand{\halftablespace}{\,}
\setlength{\tabcolsep}{1.5pt}
\begin{table*}[t]
\caption{\textbf{Quantitative comparisons on the ACID test set.} ``MV?'' indicates whether a method requires (posed) multi-view data for training. We report view synthesis results with two different types of ground truth (shown as X/Y): sequences rendered with 3D Photos~\cite{Shih3DP20} (left), and real sequences (right). KID and Style are scaled by 10 and $10^5$ respectively.
See Sec.~\ref{sec:quantitative} for descriptions of baselines. }
\begin{center}
\small
\begin{tabular}{l l|ccc | cccc}
\toprule
& & \multicolumn{3}{c|}{View Synthesis} & \multicolumn{4}{c}{View Generation}
\\
Method & MV? &  PSNR$\uparrow$ & SSIM$\uparrow$ & LPIPS$\downarrow$ &  FID$\downarrow$ & $\text{FID}_{\text{sw}}\downarrow$ & KID$\downarrow$ & Style$\downarrow$\\ 
\midrule
GFVS~\cite{rombach2021geometry}  & Yes 
& 11.3/11.9 & 0.68/0.69  & 0.33/0.34
& 109 & 117 & 0.87 & 14.6
\\
PixelSynth~\cite{rockwell2021pixelsynth} & Yes 
& 20.0/19.7 & 0.73/0.70  & 0.19/0.20
& 111 & 119 & 1.12 & 10.54
\\
SLAMP~\cite{Akan2021ICCV}  & Yes 
& - & - & -
& 114 & 138 & 1.91 & 15.2
\\
DIGAN~\cite{yu2022digan}  & Yes 
& - & - & -
& 53.4 & 57.6 & 0.43 & 5.85
\\
Liu~\etal~\cite{liu2021infinite}  & Yes
& 23.0/\textbf{21.1} & \textbf{0.83}/\textbf{0.74}  & 0.14/{0.18}
& 32.4 & 37.2 & 0.22 & 9.37
\\
Ours  & No & 
$\textbf{23.5}$/$\textbf{21.1}$ & 0.81/0.71 & $\textbf{0.10} / \textbf{0.15}$
& $\mathbf{19.3}$ & $\mathbf{25.1}$ & $\mathbf{0.11}$ & $\mathbf{5.63}$ \\
\bottomrule
\end{tabular}
\end{center}
\label{table:acid_number}  
\end{table*}

\subsection{Metrics} \label{sec:metrics}

We evaluate synthesis quality 
on two tasks that we refer to as  
\emph{short-range view synthesis} and \emph{long-range view generation}. By (short-range) view synthesis, we mean the ability to render high fidelity views near a source view,
and we report standard error metrics between predicted and ground truth views, including PSNR, SSIM and LPIPS~\cite{zhang2018perceptual}. 
Since there is no multi-view data for LHQ, we create pseudo ground truth images over a trajectory of length $5$ from a global LDI mesh~\cite{shade1998layered} computed using 3D Photos~\cite{Shih3DP20}; please see the supplemental material for more details. 
On the ACID dataset, we report errors on real video sequences where we use SfM-aligned depth maps to render images from each method. We also report results from ground truth sequences created with 3D Photos, since we observe that in real video sequences, pixel misalignments can also be caused by factors like scene motion and errors in monocular depth or camera poses.

For the task of (long-range) view generation, following prior work~\cite{liu2021infinite} we adopt the Fréchet Inception Distance (FID), sliding window FID ($\text{FID}_{\text{sw}}$) (with window size $\omega=20$), and Kernel Inception Distance (KID)~\cite{binkowski2018demystifying} to measure the synthesis quality of different methods.
We also introduce a style consistency metric that computes an average style loss between the starting image and each generated view along a camera trajectory. This metric reflects how much the style of a generated sequence deviates from the original image; we evaluate it over a trajectory of length 50. For FID and KID calculations, we compute real statistics from 50K images randomly sampled from each dataset, and calculate fake statistics from 70K and 100K generated images on ACID and LHQ respectively, where 700 and 1000 test images are used as starting images evaluated over 100 steps. %
Note that since SLAMP and DIGAN do not support camera viewpoint control, we only evaluate them on the view generation task.

\setlength{\tabcolsep}{1pt}
\begin{table*}[t]
\caption{\textbf{Ablation study on the LHQ test set.} KID and Style are scaled by 10 and $10^5$ respectively. See Sec.~\ref{sec:ablations} for a description of each baseline. }
\centering
\footnotesize	
\begin{tabular}{l | ccccc |ccc | cccc}
\toprule
& \multicolumn{5}{c|}{Configurations} &  \multicolumn{3}{c|}{View Synthesis} & \multicolumn{4}{c}{View Generation}
\\
Method 
& $\Lrecon$ & $\Ladv$ & PTG & BGS & Sky 
& PSNR$\uparrow$ & SSIM$\uparrow$ & LPIPS$\downarrow$ &  FID $\downarrow$ & $\text{FID}_{\text{sw}}\downarrow$ & KID $\downarrow$ & Style$\downarrow$
\\
\midrule
Naive
& \checkmark & \checkmark &  & &
& 28.0 & 0.87  & 0.07
& 38.1 & 52.1 & 0.25 & 6.36
\\
w/o BGS 
& \checkmark & \checkmark & \checkmark & & \checkmark
& 28.0 & 0.89  & 0.08
& 34.9 & 41.1 & 0.20 & 6.45 
\\
w/o PTG
& \checkmark & \checkmark &  & \checkmark & \checkmark 
& 28.1 & 0.90  & 0.07
& 35.3 & 42.6 & 0.21 & 6.04
\\
w/o repeat
& \checkmark &  & & & \checkmark
& 26.8 & 0.86  & 0.15
& 61.3 & 85.5 & 0.40  & 8.15
\\
w/o SVS
& & \checkmark& \checkmark & \checkmark & \checkmark 
& 26.6 & 0.85  & 0.08
& 23.4 & 30.2 & 0.12 & 6.37
\\
w/o sky
& \checkmark & \checkmark & \checkmark   & \checkmark &   
& 28.3 & 0.90  & 0.07
& 24.8 & 31.3 & 0.11 & 6.43
\\
Ours (full)  
& \checkmark   & \checkmark  & \checkmark   & \checkmark  & \checkmark
& $\mathbf{28.4}$ & $\mathbf{0.91}$ & $\mathbf{0.06}$ 
& $\mathbf{19.4}$ & $\mathbf{25.8}$ & $\mathbf{0.09}$ & $\mathbf{5.91}$ 
\\
\bottomrule
\end{tabular}
\label{table:lhq_number}  
\end{table*}

\subsection{Implementation details}
\label{sec:impl_detail}
We set the maximum camera trajectory length $\Tmax=10$. The weight of $R_1$ regularization $\lambda_2$ is set to $0.15$ and $0.004$ for the LHQ and ACID datasets, respectively. During training, we found that treating a predicted view along a long virtual trajectory as ground truth and adding a small self-supervised view synthesis loss over these predictions yields more stable view generation results. 
Therefore we set the reconstruction weight $\lambda_1=1$ for the input training image at the starting viewpoint, and $\lambda_1=0.05$ for  frames predicted on a camera trajectory.
Following~\cite{karras2020analyzing}, we apply lazy regularization to the discriminator gradient regularization every 16 training steps and adopt gradient clipping and exponential moving averaging to update the parameters of the refinement network. 
For all experiments, we train on centrally cropped images of size $128 \times 128$ for 1.8M steps with batch size 32 using 8 NVIDIA A100 GPUs, which takes $\sim$6 days to converge. 
During rendering, we use softmax splatting~\cite{niklaus2020softmax} to 3D render images via their depth maps.
Our method can also generate higher resolution $512\times512$ views. Rather than directly training the model at high resolution, which would take an estimate of 3 weeks, we train an extra super-resolution module that takes one day to converge using the same self-supervised learning idea. We refer readers to the supplementary material for more details and high-resolution results.

\begin{figure}[t]
    \centering
    \includegraphics[width=\linewidth]{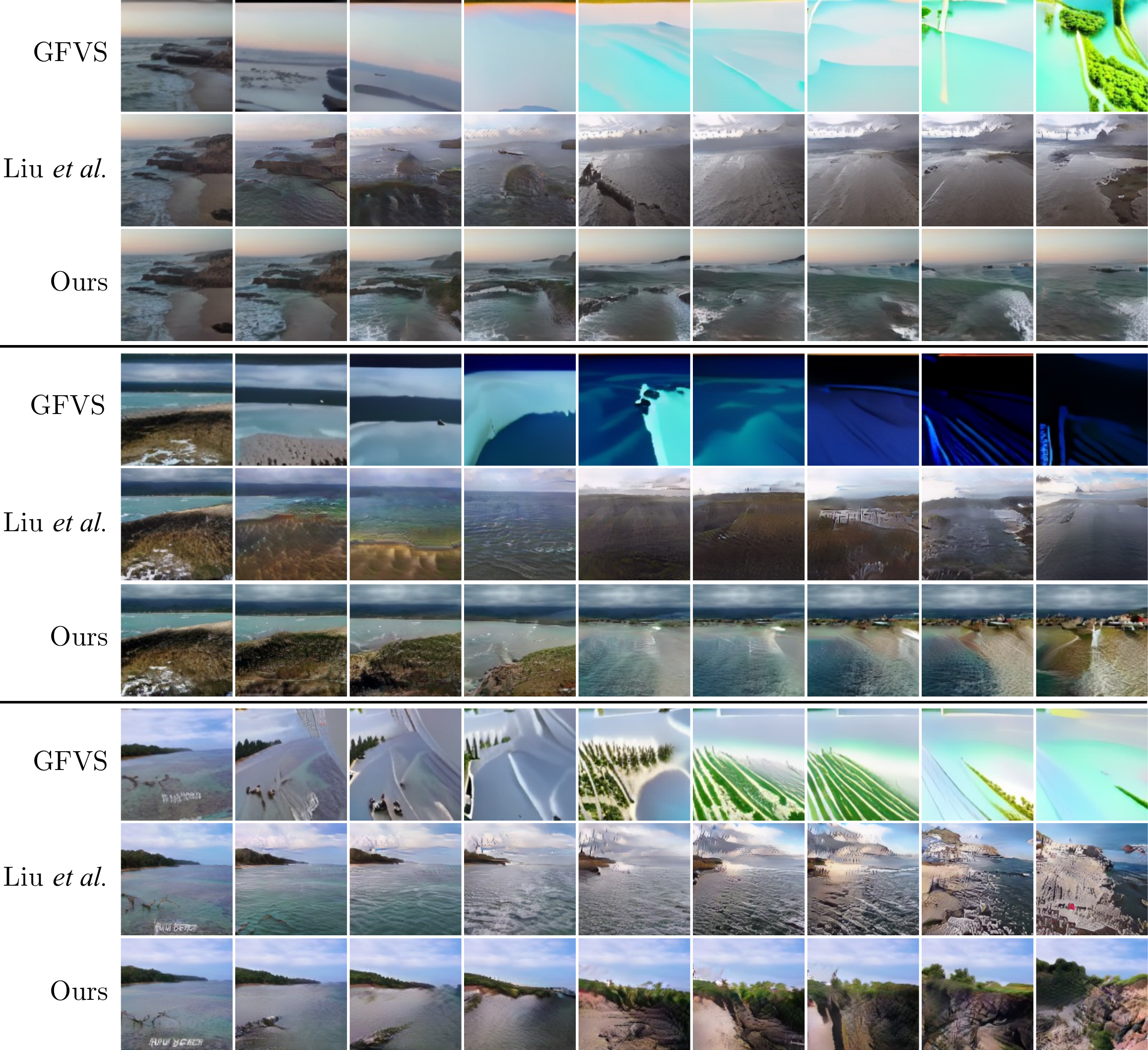} %
    \caption{\textbf{Qualitative comparisons on the ACID test set.} From left to right, we show generated views over trajectories of length 100 for three methods:
    GFVS~\cite{rombach2021geometry}, Liu~\etal~\cite{liu2021infinite} and Ours. }
    \label{fig:acid_qualitative}
\end{figure}

\subsection{Quantitative comparisons}
\label{sec:quantitative}
Table~\ref{table:acid_number} shows quantitative comparisons between our approach and other baselines on the ACID test set. Although the model only observes single images, our approach outperforms the other baselines in view generation on all error metrics, while achieving competitive performance on the view synthesis task. Specifically, our approach demonstrates the best FID and KID scores, indicating better realism and diversity for our generated views. Our method also achieves the best style consistency score.
For the view synthesis task, we achieve the best LPIPS score over the baselines, suggesting higher perceptual quality for our rendered images. We also obtain PSNR and SSIM errors on the ACID test set that are competitive with the supervised learning method from Infinite Nature, which uses a supervised reconstruction loss computed on real sequences. 

\begin{figure}[t]
    \centering
    \includegraphics[width=\linewidth]{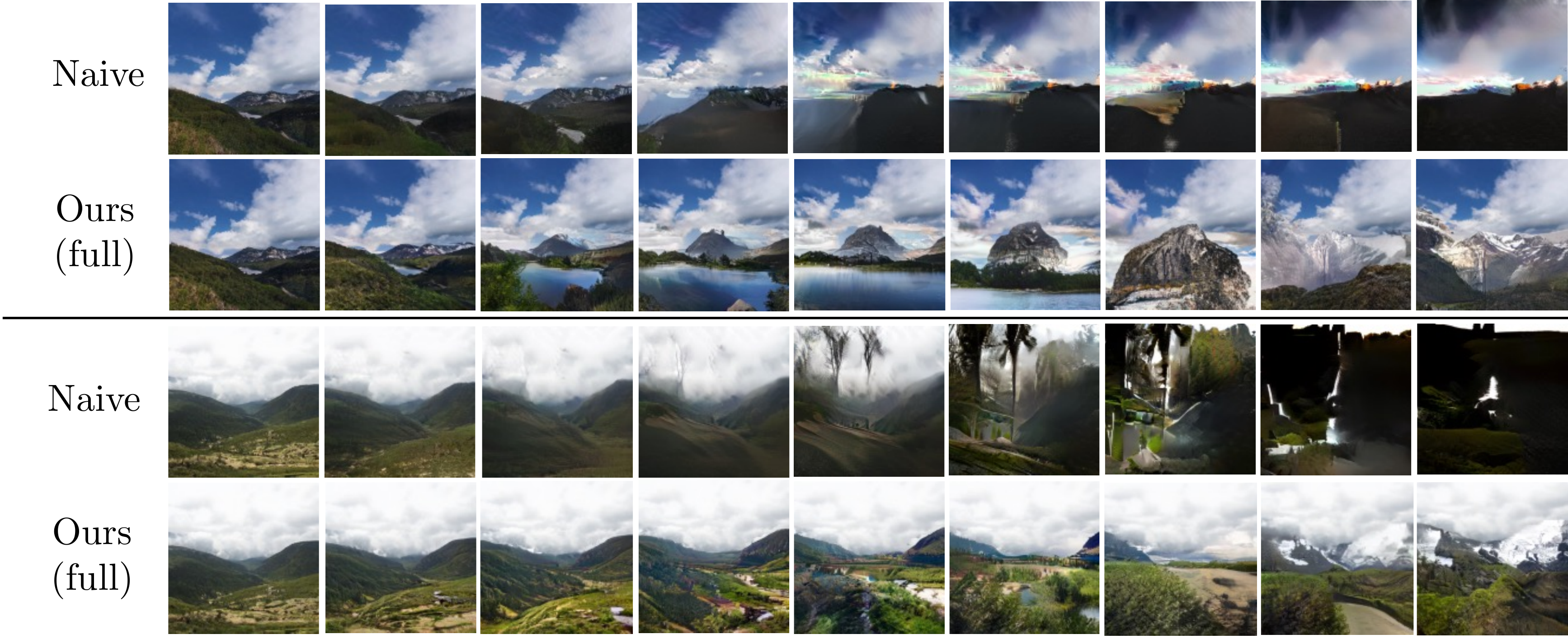}   
    \caption{\textbf{Qualitative comparisons on the LHQ test set.} On two starting views, from left to right, we show generated views over trajectories of length 100 from a naive baseline and our full approach. See Sec.~\ref{sec:ablations} for more details.}
    \label{fig:lhq_qualitative}
\end{figure}

\subsection{Ablation study} \label{sec:ablations}
We perform an ablation study on the LHQ test set to analyze the effectiveness of each component in our proposed system. 
We test the following configurations:
(1) a naive baseline where we apply an adversarial loss between all the predictions along a camera trajectory and a set of randomly sampled real photos, and apply geometry re-grounding as introduced in Infinite Nature~\cite{liu2021infinite} at test time (Naive); 
and configurations without: (2) using balanced GAN sampling (w/o BGS), (3) progressive trajectory growing (w/o PTG), (4) GAN training via long camera trajectories (w/o repeat), (5) applying self-supervised view synthesis (w/o SVS), and (6) employing global sky correction (w/o sky).
Quantitative and qualitative comparisons are shown in Table~\ref{table:lhq_number} and Fig.~\ref{fig:ablation} respectively. Our full system achieves the best view synthesis and view generation performance of these configurations. 
In particular, adding self-supervised view synthesis significantly improves view synthesis performance. Training via virtual camera trajectories, adopting introduced GAN sampling/training strategies, and applying global sky correction all improve view generation performance by a large margin.

\subsection{Qualitative comparisons}

Fig.~\ref{fig:acid_qualitative} shows visual comparisons between our approach, Infinite Nature~\cite{liu2021infinite}, and GFVS~\cite{rombach2021geometry} on the ACID test set.
GFVS quickly degenerates due to the large distance between the input and generated viewpoints. Infinite Nature
can generate plausible views over multiple steps, but the content and style of generated views quickly diverge into 
an unrelated unimodal scene. Our approach, in contrast, not only generates more consistent views with respect to starting images, but also demonstrates significantly improved synthesis quality and realism.

Fig.~\ref{fig:lhq_qualitative} shows visual comparisons between the naive baseline described in Sec.~\ref{sec:ablations} and our full approach.
The generated views from the baseline quickly deviate from realism due to ineffective training/inference strategies. In contrast, our full approach can generate much more realistic, consistent, and diverse results over long camera trajectories. For example, the views generated by our approach cover diverse and realistic natural elements such as lakes, trees, and mountains.

\subsection{Single-image perpetual view generation} 
\begin{figure}[t]
    \centering
    \setlength{\tabcolsep}{0.02cm}
    \renewcommand{\arraystretch}{0.5}
    \begin{tabular}{ccccccccc}
        \\
        \includegraphics[width=0.12\columnwidth]{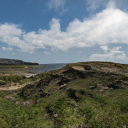} & 
        \includegraphics[width=0.12\columnwidth]{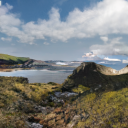} & 
        \includegraphics[width=0.12\columnwidth]{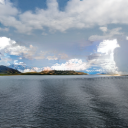} & 
        \includegraphics[width=0.12\columnwidth]{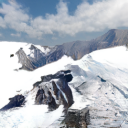} & 
        \includegraphics[width=0.12\columnwidth]{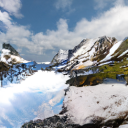} & 
        \includegraphics[width=0.12\columnwidth]{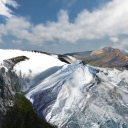} & 
        \includegraphics[width=0.12\columnwidth]{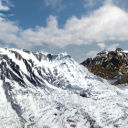} & 
        \includegraphics[width=0.12\columnwidth]{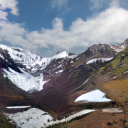} 
        \\
        \includegraphics[width=0.12\columnwidth]{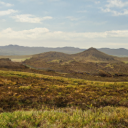} & 
        \includegraphics[width=0.12\columnwidth]{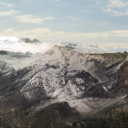} & 
        \includegraphics[width=0.12\columnwidth]{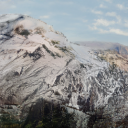} & 
        \includegraphics[width=0.12\columnwidth]{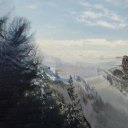} & 
        \includegraphics[width=0.12\columnwidth]{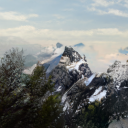} & 
        \includegraphics[width=0.12\columnwidth]{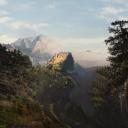} & 
        \includegraphics[width=0.12\columnwidth]{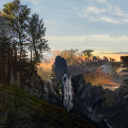} & 
        \includegraphics[width=0.12\columnwidth]{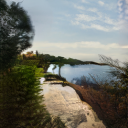} 
        \\
        \vphantom{I}\footnotesize t=0 &
        \vphantom{I}\footnotesize t=50 &
        \vphantom{I}\footnotesize t=100 &
        \vphantom{I}\footnotesize t=200 &
        \vphantom{I}\footnotesize t=250 &
        \vphantom{I}\footnotesize t=300 &
        \vphantom{I}\footnotesize t=400 &
        \vphantom{I}\footnotesize t=500 
        \\
    \end{tabular} %
    \caption{\textbf{Perpetual view generation.} Given a single RGB image, we show the results of our method generating sequences of 500 realistic new views of natural scenes without suffering significant drift. Please see video for animated results.}
    \label{fig:long_camera}
\end{figure}

Finally, we visualize our model's ability to generate long view trajectories from a single RGB image in Fig.~\ref{fig:long_camera}. Although our approach only sees single images during training, it learns to generate long sequences of 500 new views depicting realistic natural landscapes, without suffering significant drift or degeneration. We refer readers to the supplemental video for the full effect and to see results generated from different types of camera trajectories.

\section{Discussion}
\noindent\textbf{Limitations and future directions.} 
Our method inherits some limitations from prior video and view generation methods. For example, although our method produces globally consistent background sky, it does not ensure global consistency of foreground content. Addressing this issue potentially requires generating an entire 3D world model, which is an exciting future direction to explore. In addition, as with Infinite Nature, our method can generate unrealistic views if the desired camera trajectory is not seen during training (e.g.,  in-place rotation). Alternative generative methods such as VQ-VAE~\cite{razavi2019generating} and diffusion models~\cite{ho2020denoising} may provide promising paths towards addressing this limitation.

\medskip
\noindent\textbf{Conclusion.}
We presented a method for learning perpetual view generation of natural scenes 
solely from single-view photos, without requiring camera poses and multi-view data. At test time, given a single RGB image, our approach allows for generating hundreds of new views covering realistic natural scenes along a long camera trajectory. We conduct extensive experiments and demonstrate the improved performance and synthesis quality of our approach over prior supervised approaches.
We hope this work demonstrates a new step towards unbounded generative view synthesis from Internet photo collections.

\clearpage
\bibliographystyle{splncs04}
\bibliography{refs}
\end{document}